\documentclass[10pt,twocolumn,letterpaper]{article}

\usepackage{cvpr}
\usepackage{times}
\usepackage{epsfig}
\usepackage{graphicx}
\usepackage{amsmath}
\usepackage{amssymb}
\usepackage{xcolor}
\usepackage{booktabs, tabularx}
\newcommand{\ra}[1]{\renewcommand{\arraystretch}{#1}}
\usepackage{multirow}
\newcolumntype{L}[1]{>{\raggedright\arraybackslash}p{#1}}
\newcolumntype{C}[1]{>{\centering\arraybackslash}p{#1}}
\newcolumntype{R}[1]{>{\raggedleft\arraybackslash}p{#1}}
\usepackage{authblk}

\usepackage[pagebackref=true,breaklinks=true,letterpaper=true,colorlinks,bookmarks=false]{hyperref}

\cvprfinalcopy 

\def\cvprPaperID{14} 

\ifcvprfinal\fi
\author[1]{Ziyu Jiang}
\author[2]{Kate Von Ness}
\author[2]{Julie Loisel}
\author[1]{Zhangyang Wang\vspace{-0.8em}}
\affil[ ]{\textit {\{jiangziyu, k\_vonn, julieloisel, atlaswang\}@tamu.edu}}
\affil[1]{Department of Computer Science and Engineering, Texas A\&M University}
\affil[2]{Department of Geography, Texas A\&M University}
\affil[ ]{\small \url{https://github.com/geekJZY/arcticnet}}

\begin{document}

\title{ArcticNet: A Deep Learning Solution to Classify Arctic Wetlands}


\maketitle

\begin{abstract}
Arctic environments are rapidly changing under the warming climate. Of particular interest are wetlands, a type of ecosystem that constitutes the most effective terrestrial long-term carbon store. As permafrost thaws, the carbon that was locked in these wetland soils for millennia becomes available for aerobic and anaerobic decomposition, which releases carbon dioxide ($CO_2$) and methane ($CH_4$), respectively, back to the atmosphere. As $CO_2$ and $CH_4$ are potent greenhouse gases, this transfer of carbon from the land to the atmosphere further contributes to global warming, thereby increasing the rate of permafrost degradation in a positive feedback loop. Therefore, monitoring Arctic wetland health and dynamics is a key scientific task that is also of importance for policy. However, the identification and delineation of these important wetland ecosystems, remain incomplete and often inaccurate.
\vspace{-0.1em}

Mapping the extent of Arctic wetlands remains a challenge for the scientific community. Conventional, coarser remote sensing methods are inadequate at distinguishing the diverse and micro-topographically complex non-vascular vegetation that characterize Arctic wetlands, presenting the need for better identification methods. To tackle this challenging problem, we constructed and annotated the first-of-its-kind Arctic Wetland Dataset (\textbf{AWD}). Based on that, we present \textbf{ArcticNet}, a deep neural network that exploits the multi-spectral, high-resolution imagery captured from nanosatellites (Planet Dove CubeSats) with additional Digital Elevation Model (DEM) from the ArcticDEM project, to semantically label a Arctic study area into six types, in which three Arctic wetland functional types are included. We present multi-fold efforts to handle the arising challenges, including class imbalance, and the choice of fusion strategies. Preliminary results endorse the high promise of ArcticNet, achieving 93.12\% in labelling a hold-out set of regions in our Arctic study area.
\end{abstract}

\begin{figure*}
\includegraphics[scale=0.243]{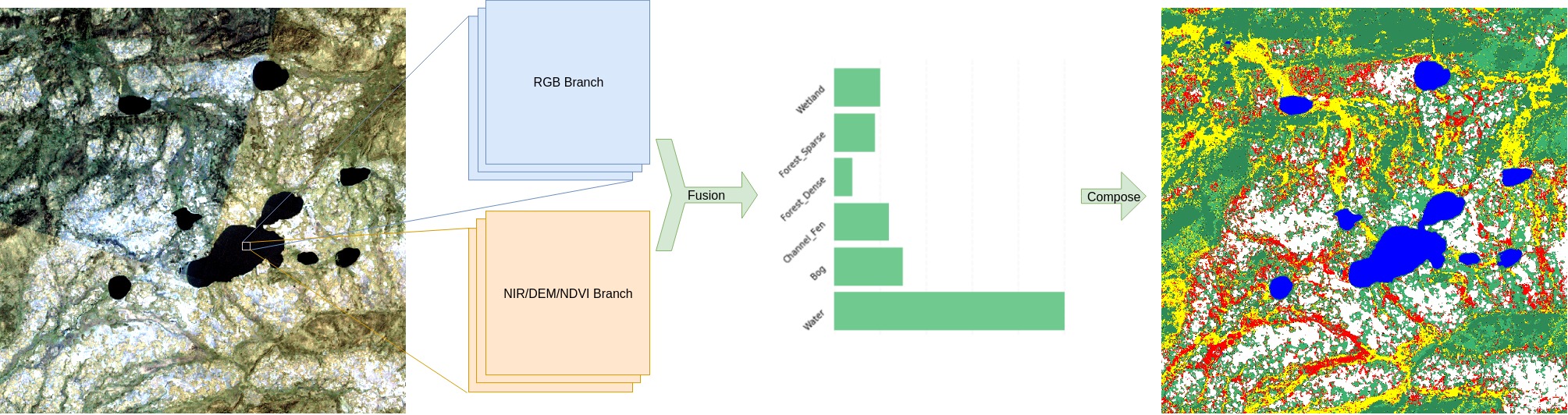}
\centering
\caption{Overview of our proposed ArcticNet. The RGB and NIR/DEM/NDVI branches takes their corresponding modality from a same patch. Then, the feature map output by two networks are fused for classification. The semantic label map of the entire area is composed by sliding over the full image.}
\label{fig:ArcticNet}
\end{figure*}

\section{Background and Motivation}

Over the past few decades, high-latitude environments have been undergoing fundamental structural and functional changes rapidly, as a result of rising temperatures. On land, these changes include intensified permafrost degradation, soil subsidence caused by melting of ground ice, formation of thermokarst terrain, longer and warmer growing seasons, changing fire regimes, and more \cite{jia2003greening}\cite{vaughan2013observations}\cite{chapin2005role}. Wetlands cover over $60\%$ of the Arctic biome, of which a sizeable fraction consists of peatlands \cite{kohnert2018toward}\cite{minayeva2018arctic}.
Peat soil accumulates over centennial to millennial timescales as a result of a net positive balance between plant production and peat decay. With growth surpassing decomposition, peat-accumulating wetlands (called peatlands) naturally sequester atmospheric carbon via photosynthesis. Under warming conditions, the fate of these soil carbon reservoirs has been questioned. On one hand, peatlands may benefit from this increased warmth by sequestering carbon at faster rates and migrating northward due to increasing plant production \cite{jones2010rapid}\cite{loisel2013recent}. On the other hand, increasing peat decay from overall warmer and drier conditions, which would lower water table and provoke greater oxidation may occur \cite{ise2008high}. 

Another important process worth considering is the transient response of wetlands to permafrost thaw. It has been shown that, in lowlands, water from permafrost thaw and melting of ground ice can form ponds, lakes, and recharge wetlands, making them even wetter. Under saturated conditions, anaerobic formation of $CH_4$ would increase, making wetlands strong carbon sources to the atmosphere \cite{harris2005detecting}\cite{kalacska2015estimation}. Another possibility is that peat-forming plants colonize shallow ponds that develop following thaw and ground subsidence; under these conditions, peat accumulation has been shown to be extremely fast \cite{jones20132200}, which makes these young peatlands highly effective carbon sinks. Overall, the response of Arctic wetlands to warming and the resulting impacts on the global carbon cycle is still ambiguous: wetlands may either continue to act as carbon stores, with possible diminished levels, or may become carbon sources in the decades to come \cite{wu2014climate}.

About 1700 billion tons of organic carbon is estimated to be stored in Arctic soils, double the amount presently in the atmosphere \cite{schuur2011climate}. Wetland maps that rely on remote sensing and in-situ data are beginning to surface \cite{xu2018peatmap}, but wetland location, spatial extent, and carbon reserve distribution need better characterization. To better understand wetland response to climate change, and make accurate estimates of carbon stocks and fluxes in wetlands, reliable and spatially-explicit representations of these ecosystems are needed. 

Recently, Big Data coupled with new data analytics are already engendering paradigm shifts across disciplines and disrupting how research is conducted. There have been recent breakthroughs in satellite technology that make it possible to obtain daily to sub-weekly high-resolution imagery of the entire planet. Our
pilot project explores the use of deep learning \cite{demir2017robocodes}\cite{audebert2018beyond}\cite{tian2018dense} to analyze this large volume of high spatial and high temporal resolution satellite-based images from the Arctic. Our end goal is to generate the first reliable Holarctic map of permafrost-affected ecosystems and address fundamental questions pertaining to the Arctic research.
As current preliminary work, we have performed multi-fold efforts:
\begin{itemize}
    \item We constructed the first-of-its-kind Arctic Wetland Dataset (AWD). The sample site of AWD is a 50km$^2$ area around the Scotty Creek Research Station, Northwest Territories, Canada, which has representative geospatial characteristics of Arctic wetlands from the discontinuous and sporadic permafrost regions. AWD includes high-resolution (3m),  multispectral imagery (RGB + Near Infrared (NIR)) and 2m Digital Elevation Model (DEM). We then carefully annotated 500 $30m\times 30m$ regions or 10 pixel by 10 pixel squares in 3m imagery (15 pixel by 15 pixel in 2m imagery) into six major classes: \textit{water}, \textit{peat bog}, \textit{channel fen}, \textit{dense forest}, \textit{sparse forest}, and \textit{wetland}.
    \item We designed a patch-level multi-modal deep network to adaptively fuse the RGB and NIR/DEM/NDVI\footnote{Normalized Difference Vegetation Index, extracted via Eqn. \ref{eq:NDVI}} modalities. In addition, we design an \textit{Augmentation Balancing} strategy to address the class-imbalance roadblock, which is caused by non uniform distribution of each class in the real world. Being a high performance deep learning solution to generate semantic maps for an Arctic study area, we named it \textit{ArcticNet}.
    \item We provided extensive ablation experiments of different models and fusion strategies, and observed significant gains by our progressive model improvements. A competitive accuracy of 93.12\% was obtained by our final model in a hold-out testing set. 
\end{itemize}
Despite calling our model ArcticNet to indicate our focus on Arctic vegetation mapping, the methodology is indeed broadly applicable to analyzing any high-resolution geospatial imagery. \textbf{We have open-sourced AWD\footnote{Please refer to our github repository}, our associated codes and pre-trained ArcticNet models}, with the hope that they can benefit similar efforts in Earth observation and remote sensing. 


\section{Related work}

\subsection{Deep learning in remote sensing}
Recently, deep learning has become increasingly integrated with high-resolution remotely sensed imagery \cite{ma2019hyperspectral}\cite{wang2014semisupervised}. \cite{long2015fully} proposed a deep semantic segmentation model by changing the fully connected layer into convolutional layers. This technique has been well received at many remote sensing tasks, including road extraction\cite{wang2015road}, building detection\cite{vakalopoulou2015building} and land cover classification\cite{kussul2017deep}. Ilke et al.\cite{demir2017robocodes} further developed an automatic generative algorithm to create street addresses from satellite imagery based on a deep learning road extraction technique. Due to the variety of remote sensing technologies and sensors (active vs. passive; radar, lidar, spectroradiometer, etc.), there are often multiple sources of overlapping data available for the same geographic areas. Therefore, many studies focus on the fusion of heterogeneous datasets. For example, multispectral imagery and depth data were fused using CNN features, hand-crafted features, and Conditional Random Fields (CRFs) by \cite{paisitkriangkrai2015effective}. \cite{audebert2018beyond} further investigated fusion of remote sensing imagery in an end-to-end manner. OpenStreetMap (OSM) was also fused in \cite{audebert2017joint}. Remote sensing data for one area through time can also be accessed and analyzed, making it a popular and useful data source for temporal analyses. \cite{li2013machine} used the latter to identify forest type changes over a 20-year period. It was also used to predict traffic speed with long short-term memory neural network (LSTM) \cite{ma2015long}. Other challenges in remote sensing image segmentation, such as high resolution \cite{chen2019collaborative}, and low resolution and visual quality \cite{liu2017image}, are also addressed. 

\subsection{Wetlands classification}
Remote sensing is being increasingly utilized for mapping wetlands and peatlands due to its extensive spatial coverage at low costs \cite{lees2018potential}. However, these ecosystems have presented a particular challenge to conventional remote sensing techniques due to 1) the incapability of optical and radar sensors to directly estimate peat depth, and therefore tease apart organic vs. mineral soils; 2) the difficulty to distinguish between wetland types due to similar vegetation cover, and 3) the seasonal changes in soil moisture in wetlands, which make sensors with high penetration capabilities ineffective \cite{krankina2008meeting}. To further complicate the matter, Arctic wetlands experience additional mapping challenges when vegetation is used as an identification proxy. Traditional remote sensing methods are tailored towards vascular vegetation, but a majority of Arctic wetland vegetation is highly heterogeneous, low-lying, and non-vascular, presenting complications. Additionally, Arctic vegetation has complex micro-topographies that are harder for commercial, coarser sensors to discern \cite{kalacska2015estimation}.
 
Hyperspectral sensors are starting to become more relevant in wetland studies (e.g. \cite{erudel2017criteria}\cite{liu2017examining}\cite{langford2019arctic}), but this data is generally less commercially-available to scientists and is more cost- and labor-intensive. Therefore, a focus needs to be placed on developing better methods for wetland identification using multispectral imagery. This study presents a unique set of high-resolution (3m), multispectral imagery (RGB + NIR) acquired from CubeSats. These nanosatellites have been mostly used for educational purposes and only recently have started to be applied to Earth Science missions \cite{selva2012survey}. While CubeSats have inevitable limitations compared to large-scale acquisition systems with respect to payload constraints, data storage, geolocational controls, power, propulsion, and thermal control, CubeSats have shown high success \cite{selva2012survey}. Cooley et al.\cite{cooley2019arctic} have so far been the only study to apply CubeSat imagery to Arctic research, using a time-series of imagery to track surface water area changes in lakes. Their study suggests high success, using machine learning and object-based classifications to overcome some of the data accuracy limitations, thanks to the high spatial and temporal imagery resolutions.

Several studies started utilizing the deep learning method for wetland identification and classification. Siewert et al.\cite{siewert2018high} employed random forest to map soil organic carbon content in permafrost terrain across the (sub-)Arctic regions. R{\"a}s{\"a}nen et al. \cite{rasanen2019usability} mapped areal coverage and changes in bare peat area over a peat plateau located in north-western Russia between 2007 and 2015 with random forest and achieved an F-score of 0.57. In our work, we show that by leveraging the deep learning method with the data fusion technique,  we can achieve an accuracy as high as 93.12\% in wetland classification.

\section{Method}

For developing our solution, we first introduce a new dataset named \textit{Arctic Wetland Dataset} (\textbf{AWD}). We then detail the ArcticNet model, which consists of two single modality patch-wise classification models for RGB and NIR/DEM/NDVI bands respectively, followed by a learnable fusion step. The model is illustrated in Figure \ref{fig:ArcticNet}.

\subsection{Arctic Wetland Dataset}



Our first contribution is the Arctic Wetland Dataset (AWD). The sample site of AWD is a 50km $^2$ area located around the Scotty Creek Research Station, Northwest Territories, Canada ($61.3^\circ N, 121.3^\circ W$). Scotty Creek is an intensively studied watershed when it comes to permafrost dynamics (e.g. \cite{mcclymont2013geophysical}\cite{haynes2018permafrost}), hydrology (e.g. \cite{hayashi2004hydrologic}\cite{quinton2013active}\cite{burd2018seasonal}), and vegetation (e.g. \cite{chasmer2011vegetation}\cite{chasmer2014decision}\cite{garon2016additions}\cite{merchant2017contributions}). Located in Canada's boreal region, it is an environment characterized by discontinuous to sporadic permafrost, coniferous forests, and wetlands (\cite{brown1997circum}\cite{merchant2017contributions}).

We argue that this site is reasonably representative of Arctic wetlands due to being characterized by: 1) forested permafrost peat plateaus, which are elevated areas dominated by black spruce trees with an understory of shrubs, lichen, and mosses; 2) channel (‘flow-through’) fens, which are lowland areas that act as water drainage pathways with some scattered trees, healthy sedges, and small-stature vegetation overlaying peat, and 3) peat bogs, which are blanketed by lichen and mosses, with some of the bogs having sparse dwarfed trees. However, we \textbf{do not claim} that this site suffices to train a model that can generalize to the whole Arctic area: it is merely a starting point. We are working to add  multiple different study areas and increase the dataset size. The transferability and generalizability across different sites would be further investigated.


We assembled the AWD by collecting high-resolution (3m), multispectral imagery (RGB and Near Infrared (NIR) from CubeSats), as well as a 2m Digital Elevation Model (DEM) from the ArcticDEM project by the Polar Geospatial Center for complement. We then selected 500 non-overlapping local areas (or patches) of $30m\times 30m$ (10 pixels by 10 pixels for 3m imagery and 15 pixels by 15 pixels for 2m)  from our study site. We categorized each patch into one of the six geospatial categories (as per the dominant type in this patch): \textit{water}, \textit{peat bog}, \textit{channel fen}, \textit{dense forest}, \textit{sparse forest} and \textit{wetland}. The categories of interest are \textit{peat bog}, \textit{channel fen} and \textit{wetland}, since they are the three critical types of wetlands our study aims at disentangling. The patch classification was performed by a geoscientist with expertise in Arctic vegetation. In addition, a drone flight video footage covering part of the study area was used to distinguish between the six main classes due to its high-resolution coverage over the varying ecosystems in this watershed. Furthermore, many published classifications and vegetation surveys related to this watershed have been used as a cautious reference. Moreover, a combination of false color imagery and a normalized difference vegetation index (NDVI) were used to complement the visual identification of the patches. Lastly, we i.i.d. split the dataset to create a training set of 300 patches, a validation set of 100, and a hold-out testing set of 100. All data are aligned by latitude and longitude. 

\begin{figure}[t]
\includegraphics[scale=0.4]{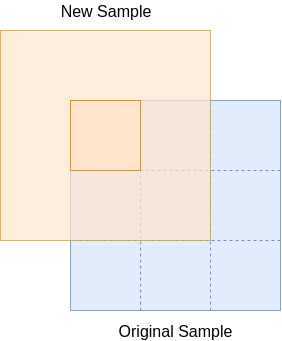}
\centering
\caption{\textit{Augmentation Balancing}. We can pick one $10m \times 10m$ area on an original sample, to be the center to crop a new sample from the imagery, sharing the same label.}
\label{fig:augmentation}
\end{figure}

\subsection{Single-Modality Backbone}


To start with, we chose two single band modality mod we employed ResNet-50 \cite{he2016deep} as the backbone patch-wise classification model for each single band modality (RGB, or NIR/DEM/NDVI band). For the RGB input, we stacked all three channels together. For the NIR/DEM/NDVI input, the Normalized Difference Vegetation Index (NDVI) band was first calculated as
\begin{equation} \label{eq:NDVI}
    \textit{NDVI} = \frac{\textit{NIR} - \textit{R}}{\textit{NIR} + \textit{R}}    
\end{equation}
Then NIR and DEM are extracted from CubeSat and ArcticDEM project, respectively, and scaled to $(0, 1)$ in order to stay within ``comparable'' value ranges with the NDVI data. Afterward, they were stacked and fed to the network. The choice of these two single modalities follows the successful practice in \cite{audebert2018beyond}, which explore the fusion of RGB with NDSM\footnote{Normalized Digital Surface Map}/DSM\footnote{Digital Surface Map}/NDVI. We replace the DSM with DEM, which is similar with it. However, NDSM band cannot be extracted from DEM. So we replace it with NIR for utilizing pretrained model which requires 3 channels' input.

\paragraph{Addressing the Class Imbalance} \label{sec:classBalance}

AWD closely follows and reflects the distribution of the six classes. However, from the perspective of model training, the class distribution is too unbalanced to ensure robust performance, especially when it comes to classifying sparse classes (such as \textit{wetland}) that are actually critical for analyzing wetlands. The class distribution in one of the training sets (since we used 3-fold cross-validation, we have multiple training sets) could be found in Table \ref{tab:trainingsetDistribution} as \textit{Original}. We propose an \textit{Augmentation Balancing} to calibrate the class balance and to enable a more robust training. 

Considering that the labeled patches are cropped from the holistic high-resolution image, a possible augmentation method could be first divide each original $30m \times 30m$ sample into nine $10m\times 10m$ sub-patches. Each sub-patch is then considered as a ``center''. One of nine ``centers'' is randomly chosen, to which a new $30m\times 30m$ sample is cropped. Finally, left-right and up-down flipping are applied to the newly-cropped $30m \times 30m$ sample in a probability of $0.5$. This augmentation strategy is illustrated in Figure \ref{fig:augmentation}. This augmentation strategy is employed for generating new samples from old ones to make all the classes the same size in terms of their number of patches (same with the largest class from the original samples). The class distribution after \textit{Augmentation Balancing} is also listed in Table \ref{tab:trainingsetDistribution} as \textit{After}.

This approach is based on two reasons: 1) We observe that the geospatial characteristic changes smoothly in those selected areas; therefore, sliding the window for a little typically will not change the label. 2) Our goal is to classify the dominant class within each patch. Therefore, introducing a small part of other class will increase the robustness instead of hurting the model. 

\begin{table}[htbp]

\caption{The distribution of the training dataset before and after \textit{Augmentation Balancing}}
{\footnotesize
\centering
\small
\resizebox{\columnwidth}{!}{
    \label{tab:trainingsetDistribution}
    \ra{1.0}
    \resizebox{\textwidth}{!}{
    \begin{tabular}{@{}lC{1.2cm}C{1.2cm}C{1.2cm}C{1.2cm}C{1.2cm}C{1.2cm}@{}} \toprule
    Phase & Water & Bog & Channel Fen & Forest Dense & Forest Sparse & Wetland \\ \midrule
    Original  & 29 & 94 & 44 & 52 & 60 & 15 \\ 
    After  & 94 & 94 & 94 & 94 &  94 & 94 \\ \bottomrule
    \end{tabular}       
    }
}
}
\end{table}


\begin{figure}[t]
\includegraphics[scale=0.36]{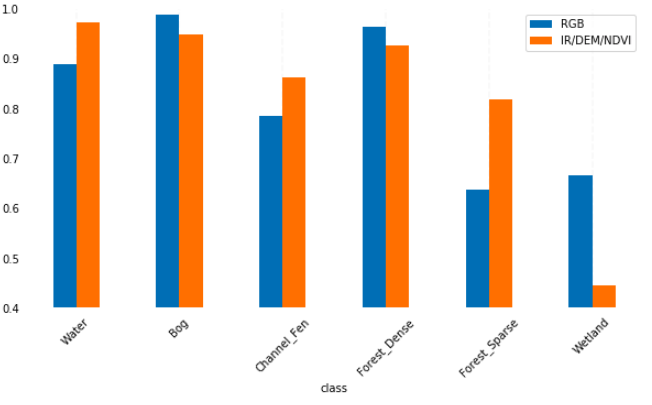}
\centering
\caption{Comparison of two single modality backbones with the accuracy on six classes.}
\label{fig:singleModailtyAcc}
\end{figure}

\subsection{Multi-Modality Fusion} \label{sec:fuse}

We compared the performance of each single modality network as shown in Figure \ref{fig:singleModailtyAcc}. Apparently, while both can classify the AWD reasonably well, there is a large discrepancy between their class-wise discriminative abilities. For example, NIR/DEM/NDVI performs better in discriminating \textit{water}, \textit{channel fen} and \textit{sparse forests}, while RGB seems to be particularly more reliable at classifying \textit{Bog}, \textit{dense forest} and \textit{wetlands}. A closer look shows that the two modalities too often yield very different classification confidences, or misaligned classification results, on individual patches. This result naturally motivated a fusion to explore their complementary powers. 

 

\begin{figure}[t]
\includegraphics[scale=0.37]{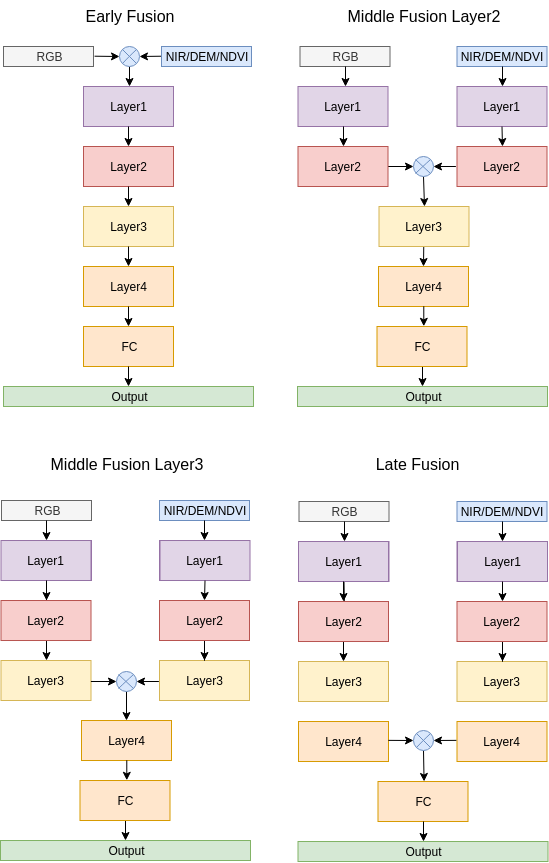}
\centering
\caption{Different fusion structures explored in our work. Middle Fusion Layer2 and Middle Fusion Layer3 are two variations of Middle Fusion. $\bigotimes$ stands for concatenation operation.}
\label{fig:fusionStructures}
\end{figure}

Concepts from deep learning-based video classification were used to explore three different fusion structures \cite{karpathy2014large}: early fusion, middle fusion and late fusion. The three fusion strategies represents different levels of flexibility control over fusing multi-modality information. Their structures are shown in Figure \ref{fig:fusionStructures}.
\begin{itemize}
    \item \textbf{Early Fusion.} RGB and NIR/DEM/NDVI modalities are directly concatenated as the network input. The resulting model will be re-trained from scratch. We observed that in early fusion, the NIR/DEM/NDVI input needs to be re-normalized to the same scale as the RGB input, to ensure stable training. 
    \item \textbf{Middle Fusion.} The feature maps produced by layer2 or layer3 of either single-modality backbone are concatenated together for the next stage joint processing (see the Middle Fusion Layer2 and Middle Fusion Layer3 in Figure \ref{fig:fusionStructures}, as two different progressive fusion ways that we tried). The resulting model will inherit all single modality model weights before the fusion, and re-train the layers after the concatenation step.
    \item \textbf{Late Fusion.} The activation vectors by layer4 of either single-modality backbone are concatenated as the input for the last fully connected layer for final classification. The resulting model will inherit all convolutional weights of both single modality models, and re-train only the last fully-connected layer.
\end{itemize}

\begin{figure*}
\includegraphics[scale=0.332]{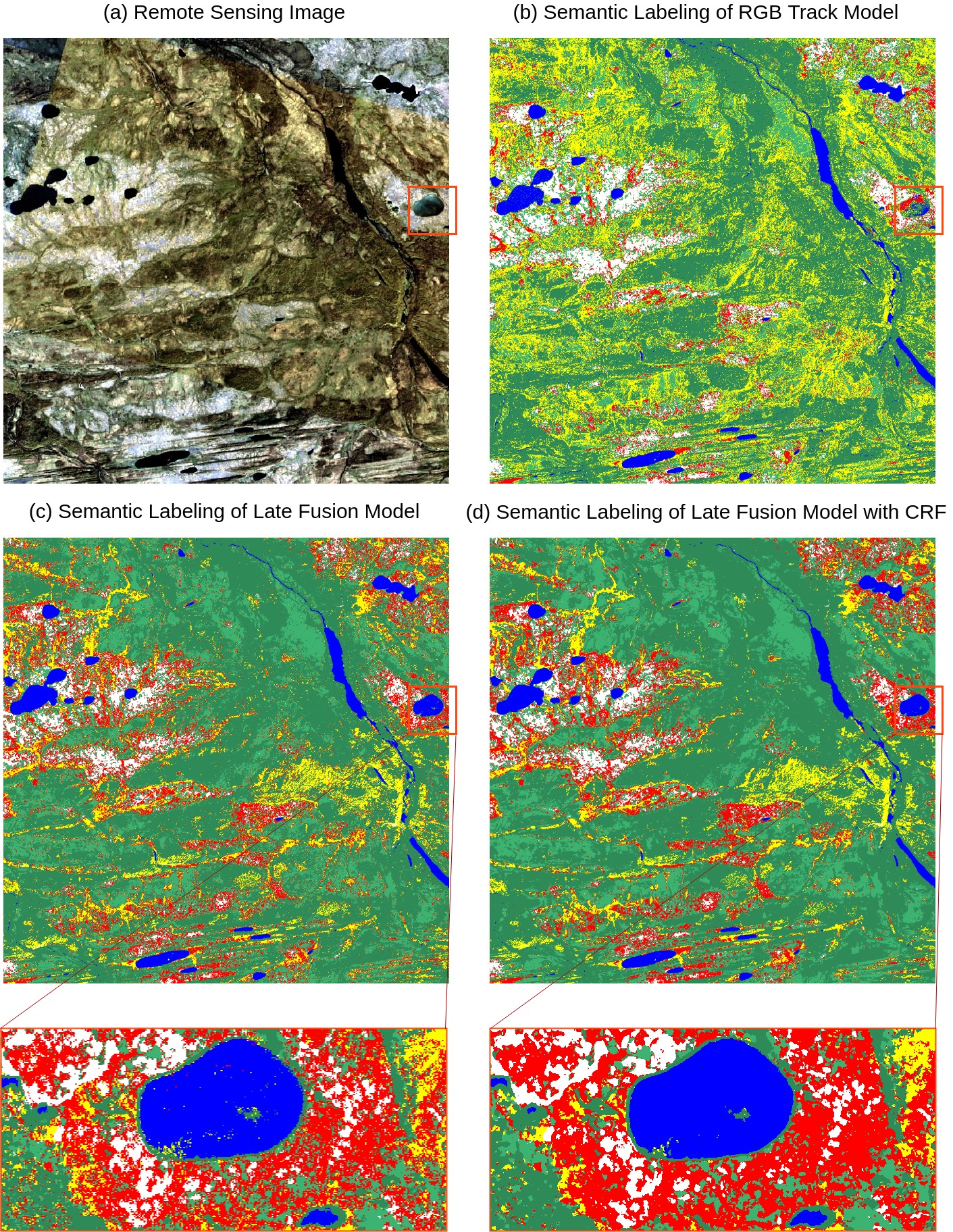}
\centering
\caption{The semantic labeling results of several variats of ArcticNet. The colors of blue, white, yellow, dark green, light green and red denote the classes of \textit{water}, \textit{bog}, \textit{channel fen}, \textit{dense forest}, \textit{sparse forest} and \textit{wetland}, respectively.}
\label{fig:experimentShow}
\end{figure*}

\begin{table*}[h]

\caption{The ablation experiment results of single modality backbones and fusion models. The first column indicates the name of each model. The first row lists the name of each class; and \textit{Overall} stands for the averaged accuracy of all classes.}
{\footnotesize
\centering
    \label{tab:arctic_ablation_results}
    \ra{1.0}
    \resizebox{\textwidth}{!}{
    \begin{tabular}{@{}lc|cccccccccccc|c@{}} \toprule
    Model &\phantom{a} & Water &\phantom{abc}& Bog &\phantom{abc}& Channel Fen &\phantom{abc}& Forest Dense &\phantom{abc}& Forest Sparse & \phantom{abc} & Wetland & \phantom{abc} & Overall \\ \midrule
    RGB Track no balance && 0.9167 && \textbf{1.0} && 0.8824 && 0.9383 &&  0.6667 && 0.1111 && 0.7525 \\ 
    RGB Track with re-weighted loss && 0.8889 && \textbf{1.0} && 0.8824 && 0.9630 && 0.6364 && 0.3333 && 0.7840 \\
    RGB Track && 0.8889 && 0.9872 && 0.7843 && 0.9630 && 0.6364 && 0.6667 && 0.8211 \\ \midrule
    NIR/DEM/NDVI Track && 0.9722 && 0.9487 && 0.8627 && 0.9259 && 0.8182 && 0.4444 && 0.8287 \\ \midrule
    Early Fusion && \textbf{1.0} && 0.9744 && 0.8627 && \textbf{1.0} && 0.8333 && 0.5556 && 0.8710 \\ 
    Middle Fusion Layer2  && \textbf{1.0} && 0.9744 && 0.8235 && 0.9877 && \textbf{0.9091} && 0.7778 && 0.9121  \\ 
    Middle Fusion Layer3 && 0.9722 && 0.9744 && \textbf{0.9020} && 0.9877 && 0.8636 && 0.5556 && 0.8758 \\
    Late Fusion && 0.9167 && 0.9615 && 0.8824 && 0.9630 && 0.8636 && \textbf{1.0} && \textbf{0.9312} \\ \bottomrule
    \end{tabular}       
    }
}
\end{table*}

\section{Experiment}

\subsection{Training setting}

We trained our models in AWD dataset in an end-to-end fashion. We used the ImageNet pre-trained ResNet-50 model as an initialization, and fine-tuned it on the AWD training set using Stochastic Gradient Descent (SGD). 
We employed the ``poly" learning rate policy described in \cite{chen2018deeplab}:
\begin{equation}
    r = r_i \times (1 - \frac{\textit{epoch}}{\textit{max epoch}})^{\textit{power}},
\end{equation}
where $r_i$ is set as $1 \times 10^{-4}$, $\textit{power}$ is set as $4$. The weight decay is set as $2 \times 10^{-5}$ and the momentum is $0.5$. 
The classification accuracy on the AWD testing set was used as the evaluation metric. A three-fold cross-validation was employed and the final accuracy is reported as the averaged result. Our model is trained on \textit{Nvidia 1080 Ti}, and it takes around $2$ hours and 2 GB GPU memory to train.


\subsection{Performance evaluation and analysis}

\textbf{A single modality model on RGB is a strong baseline.} We train the single modality backbone with only RGB images . The result is shown in Table \ref{tab:arctic_ablation_results}, referred to as as \textit{RGB Track no balance}. The overall accuracy achieves $75.25\%$, with some classes showing very high accuracy, e.g., \textit{Bog}. We consider this result to be promising, in reference to performance levels reported in similar studies \cite{rasanen2019usability}. However, the accuracy of the \textit{Wetland} class is very low (only $11.11\%$), as a result of suffering from class imbalance.


\textbf{Augmentation Balancing yields significant improvement.} When it comes to re-balancing classes, a common off-the-shelf option is to replace the standard cross entropy loss with a re-weighted loss \cite{lin2017focal}. The loss has a weight vector $\alpha = [\alpha _1, \alpha _2, \ldots, \alpha _n ]$, which is calculated by the inverse class frequencies:
\begin{equation}
    \alpha _i = \frac{\frac{1}{f_i}}{\sum_{q=0}^{n} \frac{1}{f_q}}
\end{equation}
$f_i$ represents the frequency of corresponding class. The resulting model, referred to as \textit{RGB Track with re-weighted loss} in Table \ref{tab:arctic_ablation_results}, shows overall accuracy improvement as well as on the \textit{wetland} class accuracy. We then apply the proposed \textit{Augmentation Balancing} to re-training the original model (without using re-weighted loss), and find it (called \textit{RGB Track} in Table \ref{tab:arctic_ablation_results}) to boost the performance even more than the competitive alternative of \textit{RGB Track with re-weighted loss}. That endorses the remarkable effectiveness of \textit{Augmentation Balancing} in conquering the class imbalance challenge. We therefore adopt \textit{Augmentation Balancing} by default hereinafter.


\textbf{A single modality model on NIR/DEM/NDVI performs reasonably well and provides complementary power.} We trained single modality backbone on NIR/DEM/NDVI with \textit{Augmentation Balancing}. The result was displayed in Table \ref{tab:arctic_ablation_results} as \textit{NIR/DEM/NDVI Track}. As can be observed, its class-wise accuracy distribution notably differs from that of the RGB track. While its accuracy is higher in \textit{water}, \textit{channel fen} and \textit{sparse forest}, the performance on other classes is less competitive than in RGB, especially on \textit{wetland}. Meanwhile, the overall accuracy is comparable between these two tracks.


\begin{figure}[t]
\includegraphics[scale=0.28]{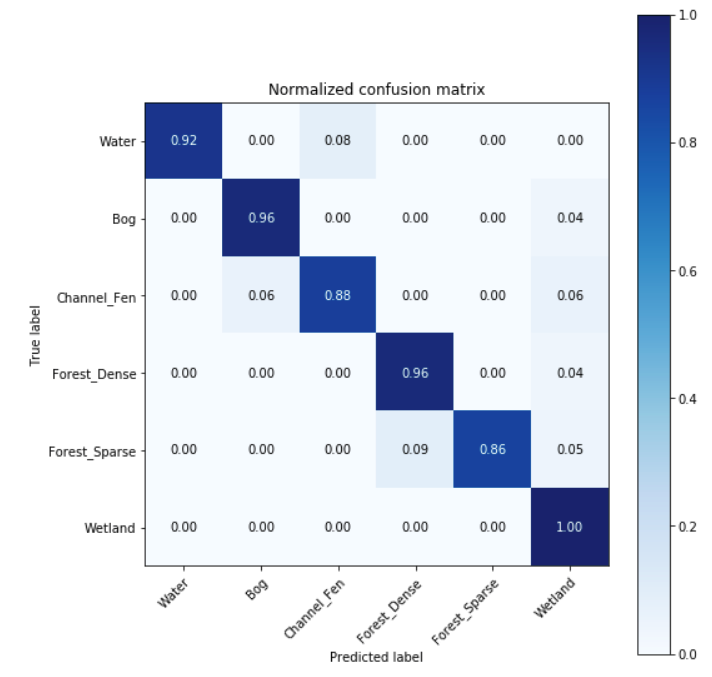}
\centering
\caption{The confusion matrix of late fusion.}
\label{fig:confusionMatric}
\end{figure}

\textbf{Fusion always helps, and Late Fusion helps the most.} We conducted experiments based on the fusion options described in Section \ref{sec:fuse}, including two variants of middle fusion. According to Table \ref{tab:arctic_ablation_results}, all fusion methods seem to improve overall classification accuracy over either single modality backone, and usually lead to more ``balanced'' class-wise accuracies, e.g., remarkably improving the \textit{wetland} class accuracy. The best model, using Late Fusion, achieves a high overall accuracy of 93.12\% and achieves perfect class-wise accuracies on \textit{wetland}.  The confusion matrix of late fusion are given in Figure \ref{fig:confusionMatric}.

Different performance shows among different fusion strategies. Although we cannot explain why it perform differently well, this is consistent with the results show in the fusion experiment in \cite{karpathy2014large}.

\begin{figure}[t]
\includegraphics[scale=0.38]{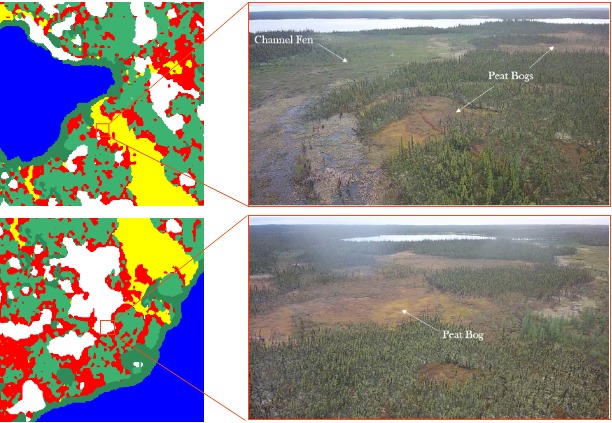}
\centering
\caption{The semantic segmentation with the drone footage in the corresponding spot.}
\label{fig:gt}
\end{figure}

\textbf{Qualitative comparison.} Figure \ref{fig:experimentShow} visualizes the semantic labeling maps. Figure \ref{fig:experimentShow}.(a) shows the visualization of raw data for target area, which contains Mosaicking caused by software visualization algorithm (Notably, we used the raw data instead of this mosaicking visualization). From Figure \ref{fig:experimentShow}.(b), which is the predicted map by the RGB Track model, we can clearly see that the lake region depicted in the red bounding box suffers from substantial misclassifications. Also, a large number of areas are mistakenly classified as \textit{(Channel Fen)}, which were verified to be incorrect by geospatial experts. By applying late fusion, those problems seem to be well alleviated. Further, we apply conditional random field (CRF) \cite{paisitkriangkrai2015effective} as a common post-processing tool in semantic segmentation to further enhance spatial consistency. Comparing the zoom in part of Figure \ref{fig:experimentShow}.(c) and \ref{fig:experimentShow}.(d), CRF alleviates the occasional image aliasing and suppresses isolated outliers. For example, the wrongly classified wetland pixels in the lake are eliminated after CRF post-processing. 
Figure \ref{fig:gt} shows some of our semantic segmentation results with the drones photos in depicted areas. From the high resolution of the drone imagery, we can identify the key vegetation differences of wetland functional types, proving the classifying results' correctness. Channel or “flow-through” fens can be identified by their lowland nature and general behavior as hydrological pathways. These ecosystems are additionally uniquely characterized by healthy sedges, short-story vegetation, and sparse trees which overlay peat. Peat bogs, on the other hand, are characterized by peat-covering lichen and moss, with some having scattered, small trees.

\textbf{Results make sense for geospatial experts.} With geospatial experts on the project, we have confirmed that the results are reasonable, primarily in the Late Fusion model. The RGB Track model highly overestimates \textit{channel fens}, when these ecosystems should only comprise about 20\% of the area. This is likely due to the fact that at 3m, it becomes difficult to distinguish \textit{channel fens} based solely on visible wavelengths that cannot capture the high amounts of healthy biomass or soil moisture that are the most defining features of these ecosystems. The Late Fusion model appears to still overestimate \textit{channel fens} in certain areas but to a much lower degree. Overall, the Late Fusion model succeeds where the RGB Track model fails by incorporating NIR information which extends the model’s ability to distinguish environments based on biophysical characteristics that the visible wavelengths cannot capture. A limitation with both models is the ability to classify other features that may be present (for example, there is a road throughout the area which is being grouped into the \textit{water} and \textit{bog} classes in the Late Fusion model). Despite this issue, all the model results are interpretable, with the Late Fusion model results the most reasonable.


\section{Conclusion}

In this work, we first introduced Arctic Wetland Dataset (AWD), a first-of-its-kind dataset sampled from a representative Arctic wetlands, with six important classes annotated. Based on this dataset, the ArcticNet was proposed for classifying imagery patches. With techniques from dataset class balancing and multi-modality fusion, our model achieves a high accuracy of 93.12\%, as well as qualitatively promising semantic label maps of the whole region. However, this work is still preliminary for mapping the whole Arctic area. In the future work, we would extend the study area and increase the dataset size. The transferability and generalizability of the model in different sites would be further investigated. Moreover, we plan explore the semi-supervised training on the AWD, to fully unleash the power of our collected (albeit unannotated) data.

Future work will also involve further analysis on different band composites (i.e. false color composite) and even different backbone neural networks that may distinguish \textit{channel fens} better. By using different variations of band combinations and backbones, we should be able to capture different types of ecosystem information throughout the study area which should help resolve model overestimation/underestimation. We will also apply these models to even higher-resolution CubeSat imagery (1m) that we have for our study area. With the higher-resolution imagery captured over the same area for a duration of 3 months, we will test how the model’s classification performs when there are distinct phenological or ecosystem changes present throughout time.


{\small
\bibliographystyle{unsrt}

}

\end{document}